\documentclass{article}

\usepackage[dblblindworkshop,final]{neurips_2026}

\usepackage[utf8]{inputenc} 
\usepackage[T1]{fontenc}    
\usepackage{hyperref}       
\usepackage{url}            
\usepackage{graphicx}
\usepackage{booktabs}
\usepackage{amsmath}
\usepackage{amssymb}
\usepackage{amsfonts}
\usepackage{nicefrac}
\usepackage{microtype}      
\usepackage{xcolor}         
\usepackage{multirow}
\title{Robust Soft Skill Adaptation via Behavioral Localization in Tool-Using Agents}
\workshoptitle{Self-Evolving Diversity-Driven Search for Robust AI Systems}

\author{%
  Zijian Jiang\\
  School of Computer Science\\
  University of Surrey\\
  Guildford, Surrey GU2 7XH, UK\\
  z.jiang@surrey.ac.uk
  \And
  Xilu Wang\\
  School of Computer Science\\
  University of Surrey\\
  Guildford, Surrey GU2 7XH, UK\\
  wangxilu@surrey.ac.uk
}

\begin{document}

\maketitle

\begin{abstract}
Text skills can extend the capabilities of tool-using agents, but long skill descriptions increase  computational costs. Soft prompts provide a compact alternative, yet learning from entire successful trajectories may also introduce unnecessary behavioral changes. We find that the useful effect of a text skill is highly concentrated in a small subset of trajectory positions. Based on this observation, we propose Behavioral Localization, which compares the frozen agent with and without the text skill to identify positions where the skill both increases the likelihood of successful tokens and meaningfully changes the next token distribution. We use these localized positions to supervise a short soft prompt. Experiments on SpreadsheetBench with a frozen Qwen3.6-35B-A3B agent show that the top 10\% of trajectory positions capture 96.12\% of the total positive skill gain. In the shared prompt setting, selective supervision improves success from 30.36\% with full trajectory to 36.79\%. Under task-specific conditions, using the top 10\% localized positions achieves 57.38\%, outperforming both the full text skill and the on-policy OPCD baseline. These results show that a small set of behaviorally relevant trajectory positions can provide effective supervision of soft skills.

\end{abstract}

\section{Introduction}
As agents are increasingly deployed across diverse domains, tasks often differ substantially in their knowledge requirements, operational procedures, and behavioral patterns. These task-specific differences are difficult to address efficiently through repeated retraining or modification of the underlying agent~\cite{AgentBench}. A flexible solution is to provide agents with task-specific guidance that helps them develop more effective task behaviors~\cite{li2026skillsbench}.

A direct approach to task adaptation is known as an agent skill, where natural language prompts are used to specify task goals, operational rules, and expected behaviors, allowing the model to perform new tasks without updating its parameters~\cite{anthropic2025agentskills}. However, this approach often relies on lengthy and carefully designed prompts, and its effectiveness can be sensitive to wording and example selection. As tasks become more complex, the prompts typically grow longer, increasing both context usage and inference cost~\cite{NEURIPS2023Gist}.

To obtain more compact and learnable skill representations, prior work has explored directly optimizing prompts in continuous representation spaces. Compared with natural language instructions, soft prompts typically require less storage and can support task adaptation by updating a small set of trainable prompt embeddings~\cite{tao2026softskill,lester-etal-2021-power}. However, these continuous vectors lack clear counterparts in natural language~\cite{deng-etal-2022-rlprompt}. As a result, while learning a target skill, a soft prompt can also introduce unexpected behavioral changes~\cite{ma-etal-2022-xprompt}. For agents, because generation is sequential, even small shifts in early predictions can propagate and amplify over subsequent steps, leading to incorrect tool selection, malformed outputs, or task execution failures~\cite{NIPS2015_e995f98d,li-etal-2023-api}. Thus, acquiring new capabilities may come at the cost of degrading abilities that the base model already possesses~\cite{zhu-etal-2022-continual}.

We find that the behavioral effect of a skill is highly concentrated in a small number of critical decision points, while much of the remaining trajectory is already correctly handled by the base agent. This suggests that robust soft skill adaptation should focus on which behaviors truly need to be learned. Based on this observation, we propose Behavioral Localization, which identifies the trajectory positions where a text skill meaningfully changes the predictions of base agent and selectively uses these positions to train a compact soft skill. At the same time, we constrain selected preservation positions toward the base distribution to reduce unnecessary behavioral drift.

Our contributions are shown below.

First, through token-level distributional analysis, we find that skill induced behavioral changes are highly sparse.

Second, we show that training a soft prompt using only selected positions by Behavioral Localization in the trajectory as supervision outperforms training on the full trajectory.

\section{Related Work}
Prompt learning provides a foundation for representing task-specific capabilities in continuous spaces. Prompt Tuning, Prefix-Tuning, and P-Tuning show that a frozen base model can be adapted by optimizing only a small set of continuous prompt embeddings~\cite{lester-etal-2021-power,li2021prefix,liu2022ptuning}. Later work explores how these representations can be reused. For example, SPoT transfers learned soft prompts from source tasks to similar target tasks~\cite{vu2022spot}. XPrompt finds that some prompt tokens can hurt downstream performance and removes these negative parts~\cite{ma-etal-2022-xprompt}. These studies mainly focus on how soft prompts are parameterized and reused. Another line of work compresses long natural language contexts into compact latent representations. Gisting compresses long prompts into a small number of reusable gist tokens~\cite{NEURIPS2023Gist}. ICAE encodes long contexts into a few memory slots that can directly condition a language model~\cite{ge2024icae}. These methods mainly aim to preserve the semantics of the original text after compression.

Agent skills differ from general text compression because their main purpose is to guide an agent's behavior. Therefore, a compressed representation should preserve the behavioral effect introduced by the skill. Recent work has begun to reuse successful agent trajectories to compress textual skills offline. SoftSkill compresses a natural-language skill into a short continuous prefix while keeping the base model frozen~\cite{tao2026softskill}. It trains the prefix by predicting the next tokens in successful trajectories. LatentSkill instead converts textual skills into task-specific LoRA adapters, moving reusable skill knowledge from the context space into the model's weight space~\cite{yu2026latentskill}. These studies show that textual agent skills can be encoded as compact latent representations. However, they mainly focus on how a skill should be represented or stored. In contrast, we study which behavioral changes introduced by a skill should be used to learn such a representation.

\section{Behavioral Locator}
\label{sec:BL}
\subsection{Problem formulation}
Let $p_{\theta}$ be a frozen agent and let $S$ denote a text skill shared by a family of tasks. Our training set consists of task description $x^i$ and it's trajectory $\tau^i=(y^{i}_1,\ldots,y^{i}_T)$. $\tau^i$ is a verified successful trajectory called the teacher trajectory. $\tau^i_t$ denotes the sequence before position $t$.
If we seek a soft prompt $P^*\in\mathbb{R}^{L\times d}$ that can replace $S$ at inference time. The parameters of the base model $\theta$ remain frozen, only $P$ is optimized. 

We aim to identify a sparse set of trajectory positions
\begin{equation}
\begin{aligned}
\mathcal{C}\subseteq{1,\ldots,T},
\qquad |\mathcal{C}|\ll T,
\end{aligned}
\end{equation}
We use $\mathcal{C}$ to supervise a short soft prompt. The goal is to focus on skill useful behavior while avoiding harmful updates induced by other trajectory positions.

\subsection{Method}
We compute the next token probability distributions along the teacher trajectory under three different conditions.
\begin{equation}
\begin{aligned}
p_{t}^{\mathrm{s}}
&= p_{\theta}(\cdot \mid S,x,\tau^i_t)\\
p_{t}^{\mathrm{n}}
&= p_{\theta}(\cdot \mid x,\tau^i_t)\\
p_{t}
&= p_{\theta}(\cdot \mid P,x,\tau^i_t)
\end{aligned}
\end{equation}
Here,$p^{\mathrm{s}}$, $p^{\mathrm{n}}$, and $p$ denote the next token distributions under the text skill, non-skill, and soft prompt, respectively. 

First, we define the positive gain $G$ induced by the skill on the base model as follows.
\begin{equation}
\begin{aligned}
\label{eq:positive_skill_gain}
G_t^i=\log p_t^{\mathrm{s}}(y_t^i)-\log p_t^{\mathrm{n}}(y_t^i)
\end{aligned}
\end{equation}
at position $t$ of the teacher trajectory, it measures how much the probability of the successful token $y_t$ increases after adding the text Skill, compared with the no-skill condition.
Next, we define the distributional shift \(JS\) induced by the skill on the base model as follows.
\begin{equation}
\begin{aligned}
\label{eq:skill_js}
JS_t^i
=
\frac{1}{2}
D_{\mathrm{KL}}
\left(
p_t^{\mathrm{s}}
\;\|\;
m_t^i
\right)
+
\frac{1}{2}
D_{\mathrm{KL}}
\left(
p_t^{\mathrm{n}}
\;\|\;
m_t^i
\right) \\
\end{aligned}
\end{equation}
where
\begin{equation}
\begin{aligned}
m_t^i
&=
\frac{1}{2}
\left(
p_t^{\mathrm{s}}
+
p_t^{\mathrm{n}}
\right)
\end{aligned}
\end{equation}
at position \(t\) of the teacher trajectory, it measures how much the full next-token distribution changes after adding the text Skill, compared with the no-skill condition. By combining the two metrics, we define the localization score $U_t^i = G_t^iJS_t^i$ at each position. We then select the top $k\%$ of positions as the training targets. 

At the selected core positions, we minimize the full vocabulary KL divergence between the predictive distributions under the text skill and soft skill conditions. At the unselected preservation positions, we use the token distribution of the no-skill model as a behavioral anchor, encouraging the soft-prompt model to preserve the original predictive behavior of the base model as much as possible.

\section{Experiment}
\begin{figure*}
    \centering
    \includegraphics[width=\linewidth]{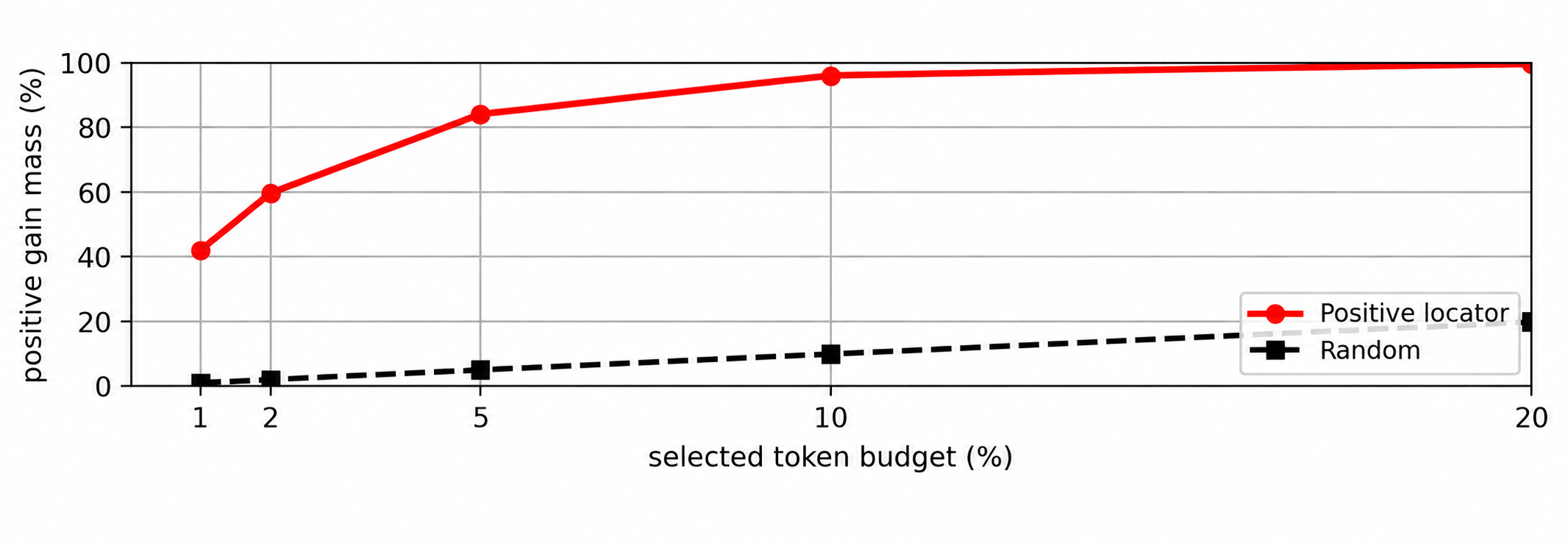}
    \caption{
        Aggregate concentration on 61 successful training trajectories.
        Positive gold token gain is highly concentrated under teacher forcing.
    }
    \label{fig:concentration}
\end{figure*}
\subsection{Setup}
We conduct our experiments on SpreadsheetBench\cite{ma2024spreadsheetbench}. We use GPT-5.5 to generate the successful teacher trajectories. All experiments use Qwen3.6-35B-A3B as the frozen student model, with all parameters fixed during training. Each soft prompt has a length of 8 and is initialized using the embeddings of the first same tokens of the full text skill.

During evaluation, we use greedy decoding with a maximum of 4,096 token cost and set the execution timeout for each task to 600 seconds. A task is considered successful only if the generated output passes all SpreadsheetBench checks.

\subsection{Main Result}

We use the Behavioral Locator to select key positions and optimize both a shared soft prompt across tasks and task-specific soft prompts for individual tasks.
As shown in Figure~\ref{fig:concentration}, the positive gain is highly concentrated in a small subset of trajectory positions. Selecting only 10\% of the tokens captures 96.12\% of the total positive Skill effect. This motivates us to use a sparse set of relevant tokens as supervision for soft skill learning.
\begin{table*}[t]
\centering
\caption{
Main SpreadsheetBench results. Bold denotes the best learned soft
prompt under each setting.
}
\label{tab:main_results}
\small
\setlength{\tabcolsep}{10pt}
\begin{tabular}{llc}
\toprule
Setting & Method & Success Rate \\
\midrule

\multirow{7}{*}{\shortstack[l]{Shared soft prompt}}
& No Skill       & 97/280 (34.64\%) \\
& Text Skill     & 118/280 (42.14\%) \\
& SoftSkill      & 85/280 (30.36\%) \\
& OPCD           & \textbf{107/280 (38.21\%)} \\

& Ours (5\%)     & 101/280 (36.07\%) \\
& Ours (10\%)    & 103/280 (36.79\%) \\

\midrule

\multirow{7}{*}{\shortstack[l]{Task-specific soft prompt}}
& No Skill                  & 26/61 (42.62\%) \\
& Text Skill                & 32/61 (52.46\%) \\
& SoftSkill                 & 28/61 (45.90\%) \\
& OPCD                      & 30/61 (49.18\%) \\

& Ours (5\%)                & 28/61 (45.90\%) \\
& Ours (10\%)               & \textbf{35/61 (57.38\%)} \\

\bottomrule
\end{tabular}
\end{table*}

No Skill measures the capability of the base agent, while Text Skill represents the up bound by providing the full text skill. SoftSkill is the offline full-trajectory compression baseline. OPCD is an important on-policy distillation baseline that trains online rather than relying on fixed teacher trajectories. Our method instead relies on lower cost offline teacher trajectories. Under the shared prompt setting, our method improves over full-trajectory SoftSkill, although it remains below the on-policy OPCD baseline. More importantly, under task-specific condition, selecting the top 10\% localized positions achieves 57.38\%, outperforming both Text Skill and OPCD. These results support a small subset of token can provide sufficient supervision for learning effective soft prompts.

\bibliographystyle{plainnat}
\bibliography{ref}

@inproceedings{AgentBench,
 author = {Liu, Xiao and Yu, Hao and Zhang, Hanchen and Xu, Yifan and Lei, Xuanyu and Lai, Hanyu and Gu, Yu and Ding, Hangliang and Men, Kaiwen and Yang, Kejuan and Zhang, Shudan and Deng, Xiang and Zeng, Aohan and Du, Zhengxiao and Zhang, Chenhui and Shen, Sheng and Zhang, Tianjun and Su, Yu and Sun, Huan and Huang, Minlie and Dong, Yuxiao and Tang, Jie},
 booktitle = {International Conference on Learning Representations},
 editor = {B. Kim and Y. Yue and S. Chaudhuri and K. Fragkiadaki and M. Khan and Y. Sun},
 pages = {52989--53046},
 title = {AgentBench: Evaluating LLMs as Agents},
 url = {https://proceedings.iclr.cc/paper_files/paper/2024/file/e9df36b21ff4ee211a8b71ee8b7e9f57-Paper-Conference.pdf},
 volume = {2024},
 year = {2024}
}

@article{li2026skillsbench,
  title={SkillsBench: Benchmarking how well agent skills work across diverse tasks},
  author={Li, Xiangyi and Liu, Yimin and Chen, Wenbo and You, Bingran and Di, Zonglin and He, Yifeng and Zheng, Shenghan and Choe, Kyoung Whan and Sun, Jiankai and Wang, Shuyi and others},
  journal={arXiv preprint arXiv:2602.12670},
  year={2026}
}

@inproceedings{NEURIPS2023Gist,
 author = {Mu, Jesse and Li, Xiang and Goodman, Noah},
 booktitle = {Advances in Neural Information Processing Systems},
 doi = {10.52202/075280-0848},
 editor = {A. Oh and T. Naumann and A. Globerson and K. Saenko and M. Hardt and S. Levine},
 pages = {19327--19352},
 publisher = {Curran Associates, Inc.},
 title = {Learning to Compress Prompts with Gist Tokens},
 url = {https://proceedings.neurips.cc/paper_files/paper/2023/file/3d77c6dcc7f143aa2154e7f4d5e22d68-Paper-Conference.pdf},
 volume = {36},
 year = {2023}
}

@misc{anthropic2025agentskills,
  author       = {{Anthropic}},
  title        = {Equipping Agents for the Real World with Agent Skills},
  year         = {2025},
  month        = oct,
  howpublished = {\url{https://www.anthropic.com/engineering/equipping-agents-for-the-real-world-with-agent-skills}},
  note         = {Anthropic Engineering Blog}
}

@misc{tao2026softskill,
      title={SoftSkill: Behavioral Compression for Contextual Adaptation}, 
      author={Xijia Tao and Yihua Teng and Xinyu Fu and Ziru Liu and Kecheng Chen and Yuzhi Zhao and Suiyun Zhang and Rui Liu and Lingpeng Kong},
      year={2026},
      eprint={2606.20333},
      archivePrefix={arXiv},
      primaryClass={cs.AI},
      url={https://arxiv.org/abs/2606.20333}, 
}

@inproceedings{lester-etal-2021-power,
    title = "The Power of Scale for Parameter-Efficient Prompt Tuning",
    author = "Lester, Brian  and
      Al-Rfou, Rami  and
      Constant, Noah",
    editor = "Moens, Marie-Francine  and
      Huang, Xuanjing  and
      Specia, Lucia  and
      Yih, Scott Wen-tau",
    booktitle = "Proceedings of the 2021 Conference on Empirical Methods in Natural Language Processing",
    month = nov,
    year = "2021",
    address = "Online and Punta Cana, Dominican Republic",
    publisher = "Association for Computational Linguistics",
    url = "https://aclanthology.org/2021.emnlp-main.243/",
    doi = "10.18653/v1/2021.emnlp-main.243",
    pages = "3045--3059",
}

@inproceedings{deng-etal-2022-rlprompt,
    title = "{RLP}rompt: Optimizing Discrete Text Prompts with Reinforcement Learning",
    author = "Deng, Mingkai  and
      Wang, Jianyu  and
      Hsieh, Cheng-Ping  and
      Wang, Yihan  and
      Guo, Han  and
      Shu, Tianmin  and
      Song, Meng  and
      Xing, Eric  and
      Hu, Zhiting",
    editor = "Goldberg, Yoav  and
      Kozareva, Zornitsa  and
      Zhang, Yue",
    booktitle = "Proceedings of the 2022 Conference on Empirical Methods in Natural Language Processing",
    month = dec,
    year = "2022",
    address = "Abu Dhabi, United Arab Emirates",
    publisher = "Association for Computational Linguistics",
    url = "https://aclanthology.org/2022.emnlp-main.222/",
    doi = "10.18653/v1/2022.emnlp-main.222",
    pages = "3369--3391",
}

@inproceedings{ma-etal-2022-xprompt,
    title = "{XP}rompt: Exploring the Extreme of Prompt Tuning",
    author = "Ma, Fang  and
      Zhang, Chen  and
      Ren, Lei  and
      Wang, Jingang  and
      Wang, Qifan  and
      Wu, Wei  and
      Quan, Xiaojun  and
      Song, Dawei",
    editor = "Goldberg, Yoav  and
      Kozareva, Zornitsa  and
      Zhang, Yue",
    booktitle = "Proceedings of the 2022 Conference on Empirical Methods in Natural Language Processing",
    month = dec,
    year = "2022",
    address = "Abu Dhabi, United Arab Emirates",
    publisher = "Association for Computational Linguistics",
    url = "https://aclanthology.org/2022.emnlp-main.758/",
    doi = "10.18653/v1/2022.emnlp-main.758",
    pages = "11033--11047",
}

@inproceedings{NIPS2015_e995f98d,
 author = {Bengio, Samy and Vinyals, Oriol and Jaitly, Navdeep and Shazeer, Noam},
 booktitle = {Advances in Neural Information Processing Systems},
 editor = {C. Cortes and N. Lawrence and D. Lee and M. Sugiyama and R. Garnett},
 pages = {},
 publisher = {Curran Associates, Inc.},
 title = {Scheduled Sampling for Sequence Prediction with Recurrent Neural Networks},
 url = {https://proceedings.neurips.cc/paper_files/paper/2015/file/e995f98d56967d946471af29d7bf99f1-Paper.pdf},
 volume = {28},
 year = {2015}
}

@inproceedings{li-etal-2023-api,
    title = "{API}-Bank: A Comprehensive Benchmark for Tool-Augmented {LLM}s",
    author = "Li, Minghao  and
      Zhao, Yingxiu  and
      Yu, Bowen  and
      Song, Feifan  and
      Li, Hangyu  and
      Yu, Haiyang  and
      Li, Zhoujun  and
      Huang, Fei  and
      Li, Yongbin",
    editor = "Bouamor, Houda  and
      Pino, Juan  and
      Bali, Kalika",
    booktitle = "Proceedings of the 2023 Conference on Empirical Methods in Natural Language Processing",
    month = dec,
    year = "2023",
    address = "Singapore",
    publisher = "Association for Computational Linguistics",
    url = "https://aclanthology.org/2023.emnlp-main.187/",
    doi = "10.18653/v1/2023.emnlp-main.187",
    pages = "3102--3116",
}

@inproceedings{zhu-etal-2022-continual,
    title = "Continual Prompt Tuning for Dialog State Tracking",
    author = "Zhu, Qi  and
      Li, Bing  and
      Mi, Fei  and
      Zhu, Xiaoyan  and
      Huang, Minlie",
    editor = "Muresan, Smaranda  and
      Nakov, Preslav  and
      Villavicencio, Aline",
    booktitle = "Proceedings of the 60th Annual Meeting of the Association for Computational Linguistics (Volume 1: Long Papers)",
    month = may,
    year = "2022",
    address = "Dublin, Ireland",
    publisher = "Association for Computational Linguistics",
    url = "https://aclanthology.org/2022.acl-long.80/",
    doi = "10.18653/v1/2022.acl-long.80",
    pages = "1124--1137",
}

@inproceedings{li2021prefix,
  title={Prefix-Tuning: Optimizing Continuous Prompts for Generation},
  author={Li, Xiang Lisa and Liang, Percy},
  booktitle={Proceedings of the 59th Annual Meeting of the Association for
  Computational Linguistics and the 11th International Joint Conference on
  Natural Language Processing (Volume 1: Long Papers)},
  pages={4582--4597},
  year={2021},
  publisher={Association for Computational Linguistics},
  doi={10.18653/v1/2021.acl-long.353},
  url={https://aclanthology.org/2021.acl-long.353/}
}

@inproceedings{liu2022ptuning,
  title={{P}-Tuning: Prompt Tuning Can Be Comparable to Fine-tuning Across
  Scales and Tasks},
  author={Liu, Xiao and Ji, Kaixuan and Fu, Yicheng and Tam, Weng and
  Du, Zhengxiao and Yang, Zhilin and Tang, Jie},
  booktitle={Proceedings of the 60th Annual Meeting of the Association for
  Computational Linguistics (Volume 2: Short Papers)},
  pages={61--68},
  year={2022},
  publisher={Association for Computational Linguistics},
  doi={10.18653/v1/2022.acl-short.8},
  url={https://aclanthology.org/2022.acl-short.8/}
}

@inproceedings{vu2022spot,
  title={{SP}o{T}: Better Frozen Model Adaptation through Soft Prompt Transfer},
  author={Vu, Tu and Lester, Brian and Constant, Noah and Al-Rfou', Rami
  and Cer, Daniel},
  booktitle={Proceedings of the 60th Annual Meeting of the Association for
  Computational Linguistics (Volume 1: Long Papers)},
  pages={5039--5059},
  year={2022},
  publisher={Association for Computational Linguistics},
  doi={10.18653/v1/2022.acl-long.346},
  url={https://aclanthology.org/2022.acl-long.346/}
}

@inproceedings{ge2024icae,
  title={In-context Autoencoder for Context Compression in a Large Language Model},
  author={Ge, Tao and Jing, Hu and Wang, Lei and Wang, Xun and Chen, Si-Qing
  and Wei, Furu},
  booktitle={The Twelfth International Conference on Learning Representations},
  year={2024},
  url={https://openreview.net/forum?id=uREj4ZuGJE}
}

@misc{yu2026latentskill,
  title={LatentSkill: From In-Context Textual Skills to In-Weight Latent Skills
  for LLM Agents},
  author={Yu, Aofan and Zhou, Chenyu and Xu, Tianyi and Guo, Zihan and
  Shan, Rong and Fu, Zhihui and Wang, Jun and Liu, Weiwen and Yu, Yong and
  Zhang, Weinan and Lin, Jianghao},
  year={2026},
  eprint={2606.06087},
  archivePrefix={arXiv},
  primaryClass={cs.AI},
  url={https://arxiv.org/abs/2606.06087}
}

@article{ma2024spreadsheetbench,
  title={Spreadsheetbench: Towards challenging real world spreadsheet manipulation},
  author={Ma, Zeyao and Zhang, Bohan and Zhang, Jing and Yu, Jifan and Zhang, Xiaokang and Zhang, Xiaohan and Luo, Sijia and Wang, Xi and Tang, Jie},
  journal={Advances in Neural Information Processing Systems},
  volume={37},
  pages={94871--94908},
  year={2024}
}
\end{document}